\documentclass{article} 
\usepackage{iclr2017_conference,times}
\usepackage{hyperref}
\usepackage{url}
\usepackage{graphicx}

\usepackage{amsmath}
\usepackage{stmaryrd}
\usepackage{dsfont}
\usepackage{algorithm}
\usepackage{algorithmic}

\usepackage{booktabs}
\usepackage{footnote}
\makesavenoteenv{tabular}
\makesavenoteenv{table}

\usepackage[titletoc]{appendix}

\iclrfinalcopy
\title{Hadamard Product \\for Low-rank Bilinear Pooling}

\author{Jin-Hwa Kim\\
Interdisciplinary Program in Cognitive Science\\
Seoul National University\\
Seoul 08826, Republic of Korea \\
\texttt{jhkim@bi.snu.ac.kr} \\
\And
Kyoung-Woon On\\
School of Computer Science and Engineering\\
Seoul National University\\
Seoul 08826, Republic of Korea \\
\texttt{kwon@bi.snu.ac.kr} \\
\And
Woosang Lim\\
School of Computing, KAIST\\
Daejeon 34141, Republic of Korea \\
\texttt{quasar17@kaist.ac.kr} \\
\And
Jeonghee Kim \& Jung-Woo Ha \\
NAVER LABS Corp. \& NAVER Corp.\\
Gyeonggi-do 13561, Republic of Korea \\
\texttt{\{jeonghee.kim,jungwoo.ha\}@navercorp.com} \\
\AND
Byoung-Tak Zhang\\
School of Computer Science and Engineering \& Interdisciplinary Program in Cognitive Science \\
Seoul National University \& Surromind Robotics\\
Seoul 08826, Republic of Korea \\
\texttt{btzhang@bi.snu.ac.kr}
}

\newcommand{\vx}[0]{\mathbf{x}}
\newcommand{\vy}[0]{\mathbf{y}}
\newcommand{\vv}[0]{\mathbf{v}}
\newcommand{\vw}[0]{\mathbf{w}}
\newcommand{\vf}[0]{\mathbf{f}}
\newcommand{\vq}[0]{\mathbf{q}}

\newcommand{\vb}[0]{\mathbf{b}}

\newcommand{\vs}[0]{\mathbf{s}}
\newcommand{\vh}[0]{\mathbf{h}}

\newcommand{\vz}[0]{\mathbf{z}}

\newcommand{\mU}[0]{\mathbf{U}}
\newcommand{\mV}[0]{\mathbf{V}}
\newcommand{\mW}[0]{\mathbf{W}}
\newcommand{\mP}[0]{\mathbf{P}}
\newcommand{\mF}[0]{\mathbf{F}}

\newcommand{\R}[0]{\mathds{R}}
\newcommand{\Z}[0]{\mathds{Z}}
\newcommand{\N}[0]{\mathds{N}}

\usepackage{xspace}
\newcommand{\eg}[0]{\textit{e.g.}}
\newcommand{\0}[0]{\hspace{0.5em}}

\DeclareMathOperator*{\argmax}{arg\,max}

\newcommand{\softmax}[0]{\text{softmax}}
\newcommand{\diag}[0]{\text{diag}}
\newcommand{\FFT}[0]{\text{FFT}}

\begin{document}

\maketitle

\begin{abstract}
Bilinear models provide rich representations compared with linear models. They have been applied in various visual tasks, such as object recognition, segmentation, and visual question-answering, to get state-of-the-art performances taking advantage of the expanded representations. However, bilinear representations tend to be high-dimensional, limiting the applicability to computationally complex tasks. We propose low-rank bilinear pooling using Hadamard product for an efficient attention mechanism of multimodal learning. We show that our model outperforms compact bilinear pooling in visual question-answering tasks with the state-of-the-art results on the VQA dataset, having a better parsimonious property.
\end{abstract}

\section{Introduction}

Bilinear models~\citep{tenenbaum2000} provide richer representations than linear models. To exploit this advantage, fully-connected layers in neural networks can be replaced with bilinear pooling. The outer product of two vectors (or Kroneker product for matrices) is involved in bilinear pooling, as a result of this, all pairwise interactions among given features are considered. Recently, a successful application of this technique is used for fine-grained visual recognition~\citep{Lin2015}.

However, bilinear pooling produces a high-dimensional feature of quadratic expansion, which may constrain a model structure and computational resources. For example, an outer product of two feature vectors, both of which have 1K-dimensionality, produces a million-dimensional feature vector. Therefore, for classification problems, the choice of the number of target classes is severely constrained, because the number of parameters for a standard linear classifier is determined by multiplication of the size of the high-dimensional feature vector and the number of target classes.

Compact bilinear pooling~\citep{Gao2015a} reduces the quadratic expansion of dimensionality by two orders of magnitude, retaining the performance of the full bilinear pooling. This approximation uses sampling-based computation, Tensor Sketch Projection~\citep{charikar2002,pham2013}, which utilizes an useful property that $\Psi(x \otimes y,h,s)=\Psi(x,h,s)*\Psi(y,h,s)$, which means the projection of outer product of two vectors is the convolution of two projected vectors. Here, $\Psi$ is the proposed projection function, and, $h$ and $s$ are randomly sampled parameters by the algorithm.


Nevertheless, compact bilinear pooling embraces two shortcomings. One comes from the sampling approach. Compact bilinear pooling relies on a favorable property, $E[\langle\Psi(x,h,s), \Psi(y,h,s)\rangle]=\langle x,y \rangle$, which provides a basis to use projected features instead of original features. Yet, calculating the exact expectation is computationally intractable, so, the random parameters, $h$ and $s$ are fixed during training and evaluation. This practical choice leads to the second. The projected dimension of compact bilinear pooling should be large enough to minimize the bias from the fixed parameters. Practical choices are 10K and 16K for 512 and 4096-dimensional inputs, respectively~\citep{Gao2015a,Fukui2016}. Though, these \textit{compacted} dimensions are reduced ones by two orders of magnitude compared with full bilinear pooling, such high-dimensional features could be a bottleneck for computationally complex models.

We propose low-rank bilinear pooling using Hadamard product (element-wise multiplication), which is commonly used in various scientific computing frameworks as one of tensor operations. The proposed method factors a three-dimensional weight tensor for bilinear pooling into three two-dimensional weight matrices, which enforces the rank of the weight tensor to be low-rank. As a result, two input feature vectors linearly projected by two weight matrices, respectively, are computed by Hadamard product, then, followed by a linear projection using the third weight matrix. For example, the projected vector $\vz$ is represented by $\mW_z^T(\mW_\vx^T\vx \circ \mW_\vy^T\vy)$, where $\circ$ denotes Hadamard product. 

We also explore to add non-linearity using non-linear activation functions into the low-rank bilinear pooling, and shortcut connections inspired by deep residual learning~\citep{He2015}. Then, we show that it becomes a simple baseline model~\citep{Antol2015} or one-learning block of Multimodal Residual Networks~\citep{Kim2016b} as a low-rank bilinear model, yet, this interpretation has not be done.

Our contributions are as follows: First, we propose low-rank bilinear pooling to approximate full bilinear pooling to substitute compact bilinear pooling. Second, Multimodal Low-rank Bilinear Attention Networks (MLB) having an efficient attention mechanism using low-rank bilinear pooling is proposed for visual question-answering tasks. MLB achieves a new state-of-the-art performance, and has a better parsimonious property. Finally, ablation studies to explore alternative choices, \eg~network depth, non-linear functions, and shortcut connections, are conducted.

\section{Low-rank Bilinear Model}

Bilinear models use a quadratic expansion of linear transformation considering every pair of features. \begin{align}
  \label{eq:bilin}
   f_i &= \sum_{j=1}^{N}\sum_{k=1}^{M}w_{ijk}x_{j}y_{k} + b_i = \vx^T\mW_i\vy + b_i
\end{align}
where $\vx$ and $\vy$ are input vectors, $\mW_i \in \R^{N \times M}$ is a weight matrix for the output $f_i$, and $b_i$ is a bias for the output $f_i$. Notice that the number of parameters is $L \times (N \times M + 1)$ including a bias vector $\vb$, where $L$ is the number of output features.

\citet{Pirsiavash2009} suggest a low-rank bilinear method to reduce the rank of the weight matrix $\mW_i$ to have less number of parameters for regularization. They rewrite the weight matrix as $\mW_i = \mU_{i}\mV_{i}^T$ where $\mU_i \in \R^{N \times d}$ and $\mV_i \in \R^{M \times d}$, which imposes a restriction on the rank of $\mW_i$ to be at most $d \leq \min(N,M)$.

Based on this idea, $f_i$ can be rewritten as follows: \begin{align} f_i &= \vx^T\mW_i\vy + b_i
       = \vx^T\mU_{i}\mV_{i}^T\vy + b_i 
       = \mathds{1}^T(\mU_{i}^T\vx \circ \mV_{i}^T\vy) + b_i
\end{align}
where $\mathds{1} \in \R^{d}$ denotes a column vector of ones, and $\circ$ denotes Hadamard product. Still, we need two third-order tensors, $\mU$ and $\mV$, for a feature vector $\vf$, whose elements are $\{f_i\}$. To reduce the order of the weight tensors by one, we replace $\mathds{1}$ with $\mP \in \R^{d \times c}$ and $b_i$ with $\vb \in \R^c$, then, redefine as $\mU \in \R^{N \times d}$ and $\mV \in \R^{M \times d}$ to get a projected feature vector $\vf \in \R^c$. Then, we get: \begin{align}
   \label{eq:lowrank}
   \vf &= \mathds{\mP}^T(\mU^T\vx \circ \mV^T\vy) + \vb
\end{align}
where $d$ and $c$ are hyperparameters to decide the dimension of joint embeddings and the output dimension of low-rank bilinear models, respectively.

\section{Low-rank Bilinear Pooling}

A low-rank bilinear model in Equation~\ref{eq:lowrank} can be implemented using two linear mappings without biases for embedding two input vectors, Hadamard product to learn joint representations in a multiplicative way, and a linear mapping with a bias to project the joint representations into an output vector for a given output dimension. Then, we use this structure as a pooling method for deep neural networks. Now, we discuss possible variations of low-rank bilinear pooling based on this model inspired by studies of neural networks. 

\subsection{Full Model}
\label{sec:full}
In Equation~\ref{eq:lowrank}, linear projections, $U$ and $V$, can have their own bias vectors. As a result, linear models for each input vectors, $\vx$ and $\vy$, are integrated in an additive form, called as \textit{full model} for linear regression in statistics:\begin{align}
   \label{eq:fullmodel}
   \vf &= \mathds{\mP}^T\big((\mU^T\vx + \vb_x) \circ (\mV^T\vy + \vb_y)\big) + \vb \nonumber \\
       &= \mathds{\mP}^T(\mU^T\vx \circ \mV^T\vy + \mU'^T\vx + \mV'^T\vy ) + \vb'.
\end{align}
Here, $\mU'^T = \diag(\vb_y) \cdot \mU^T$, $\mV'^T = \diag(\vb_x) \cdot \mV^T$, and $\vb' = \vb + \mP^T(\vb_x \circ \vb_y)$.

\subsection{Nonlinear Activation}
\label{sec:nonlinear}

Applying non-linear activation functions may help to increase representative capacity of model. The first candidate is to apply non-linear activation functions right after linear mappings for input vectors. \begin{align}
   \label{eq:non-linear}
   \vf &= \mathds{\mP}^T\big(\sigma(\mU^T\vx) \circ \sigma(\mV^T\vy)\big) + \vb
\end{align}
where $\sigma$ denotes an arbitrary non-linear activation function, which maps any real values into a finite interval, \eg~$\text{sigmoid}$ or $\tanh$. If two inputs come from different modalities, statistics of two inputs may be quite different from each other, which may result an interference. Since the gradient with respect to each input is directly dependent on the other input in Hadamard product of two inputs.

Additional applying an activation function after the Hadamard product is not appropriate, since activation functions doubly appear in calculating gradients. However, applying the activation function only after the Hadamard product would be alternative choice (We explore this option in Section~\ref{sec:experiments}) as follows: \begin{align}
   \label{eq:afterhadamard}
   \vf &= \mathds{\mP}^T \sigma \big(\mU^T\vx \circ \mV^T\vy\big) + \vb.
\end{align}
Note that using the activation function in low-rank bilinear pooling can be found in an implementation of simple baseline for the VQA dataset~\citep{Antol2015} without an interpretation of low-rank bilinear pooling. However, notably, \citet{Wu2016a} studied learning behavior of multiplicative integration in RNNs with discussions and empirical evidences.

\subsection{Shortcut Connection}

When we apply two previous techniques, full model and non-linear activation, linear models of two inputs are nested by the non-linear activation functions. To avoid this unfortunate situation, we add shortcut connections as explored in residual learning~\citep{He2015}. \begin{align}
   \label{eq:sc}
   \vf &= \mathds{\mP}^T\big(\sigma(\mU^T\vx) \circ \sigma(\mV^T\vy)\big) + h_x(\vx) + h_y(\vy) + \vb
\end{align}
where $h_x$ and $h_y$ are shortcut mappings. For linear projection, the shortcut mappings are linear mappings. Notice that this formulation is a generalized form of the one-block layered MRN~\citep{Kim2016b}. Though, the shortcut connections are not used in our proposed model, as explained in Section~\ref{sec:results}.


\section{Multimodal Low-rank Bilinear Attention Networks}
\label{sec:mlb}

In this section, we apply low-rank bilinear pooling to propose an efficient attention mechanism for visual question-answering tasks, based on the interpretation of previous section. We assumed that inputs are a question embedding vector $\vq$ and a set of visual feature vectors $\mF$ over $S \times S$ lattice space. 

\subsection{Low-rank Bilinear Pooling in Attention Mechanism}


Attention mechanism uses an attention probability distribution $\alpha$ over $S \times S$ lattice space. Here, using low-rank bilinear pooling, $\alpha$ is defined as  \begin{align}
   \label{eq:alpha}
   \alpha &= \softmax \Big(\mP_\alpha^T \big( \sigma(\mU^T_{\vq} \vq \cdot \mathds{1}^T) \circ \sigma(\mV^T_{\mF} \mF^T) \big) \Big)
\end{align}
where $\alpha \in \R^{G \times S^2}$, $\mP_\alpha \in \R^{d \times G}$, $\sigma$ is a hyperbolic tangent function, $\mU_\vq \in \R^{N \times d}$, $\vq \in \R^N$, $\mathds{1} \in \R^{S^2}$, $\mV_\mF \in \R^{M \times d}$, and $\mF \in \R^{S^2 \times M}$. If $G > 1$, multiple glimpses are explicitly expressed as in \citet{Fukui2016}, conceptually similar to \citet{Jaderberg2015}. And, the $\softmax$ function applies to each row vector of $\alpha$. The bias terms are omitted for simplicity. 

\subsection{Multimodal Low-rank Bilinear Attention Networks}

Attended visual feature $\hat{\vv}$ is a linear combination of $\mF_i$ with coefficients $\alpha_{g,i}$. Each attention probability distribution $\alpha_{g}$ is for a glimpse $g$. For $G > 1$, $\hat{\vv}$ is the concatenation of resulting vectors $\hat{\vv}_g$ as \begin{align} \label{eq:concat}
  \hat{\vv} &= \bigparallel_{g=1}^G \sum_{s=1}^{S^2} \alpha_{g,s} \mF_{s}
\end{align}
where $\bigparallel$ denotes concatenation of vectors. The posterior probability distribution is an output of a $\softmax$ function, whose input is the result of another low-rank bilinear pooling of $\vq$ and $\hat{\vv}$ as \begin{align}
   \label{eq:output}
   p(a|\vq,\mF;\Theta) &= \softmax \Big( \mathds{\mP}_o^T \big( \sigma(\mW_\vq^T\vq) \circ \sigma(\mV_{\hat{\vv}}^T\hat{\vv}) \big) \Big) \\
  \hat{a} &= \argmax_{a \in \Omega} p(a|\vq,\mF;\Theta)
\end{align}
where $\hat{a}$ denotes a predicted answer, $\Omega$ is a set of candidate answers and $\Theta$ is an aggregation of entire model parameters.

\section{Experiments}
\label{sec:experiments}

\begin{table}[h]
\caption{The accuracies of our experimental model, Multimodal Attention Residual Networks (MARN), with respect to the number of learning blocks (L\#), the number of glimpse (G\#), the position of activation functions ($\tanh$), answer sampling, shortcut connections, and data augmentation using Visual Genome dataset, for VQA \textit{test-dev} split and Open-Ended task. Note that our proposed model, Multimodal Low-rank Bilinear Attention Networks (MLB) have no shortcut connections, compared with MARN. \textbf{MODEL}: model name, \textbf{SIZE}: number of parameters, \textbf{ALL}: overall accuracy in percentage, \textbf{Y/N}: yes/no, \textbf{NUM}: numbers, and \textbf{ETC}: others. Since \citet{Fukui2016} only report the accuracy of the ensemble model on the \textit{test-standard}, the \textit{test-dev} results of their single models are included in the last sector. Some figures have different precisions which are rounded. $*$ indicates the selected model for each experiment.}
\label{tab:results}
\begin{center}
\begin{tabular}{rrcccc}
{\bf MODEL} & {\bf SIZE} & {\bf ALL} & {\bf Y/N} & {\bf NUM} & {\bf ETC} 
\\ \hline \\
MRN-L3 & 65.0M & 61.68 & 82.28 & 38.82 & 49.25 \\
MARN-L3 & 65.5M & 62.37 & 82.31 & 38.06 & 50.83 \\
MARN-L2 & 56.3M & 63.92 & 82.88 & 37.98 & 53.59 \\
* MARN-L1 & 47.0M & 63.79 & 82.73 & 37.92 & 53.46 \\
\midrule[0.4pt]
MARN-L1-G1 & 47.0M & 63.79 & 82.73 & 37.92 & 53.46 \\
* MARN-L1-G2 & 57.7M & 64.53 & 83.41 & 37.82 & 54.43 \\
MARN-L1-G4 & 78.9M & 64.61 & 83.72 & 37.86 & 54.33 \\
\midrule[0.4pt]
No Tanh        & 57.7M & 63.58 & 83.18 & 37.23 & 52.79 \\ 
* Before-Product & 57.7M & 64.53 & 83.41 & 37.82 & 54.43 \\
After-Product  & 57.7M & 64.53 & 83.53 & 37.06 & 54.50 \\
\midrule[0.4pt]
Mode Answer    & 57.7M & 64.53 & 83.41 & 37.82 & 54.43 \\
* Sampled Answer & 57.7M & 64.80 & 83.59 & 38.38 & 54.73 \\
\midrule[0.4pt]
Shortcut    & 57.7M & 64.80 & 83.59 & 38.38 & 54.73 \\
* No Shortcut & 51.9M & 65.08 & 84.14 & 38.21 & 54.87 \\
\midrule[0.4pt]
MLB    & 51.9M & 65.08 & 84.14 & 38.21 & 54.87 \\
MLB+VG & 51.9M & 65.84 & 83.87 & 37.87 & 56.76 \\
\midrule[0.4pt]
MCB+Att~\citep{Fukui2016} & 69.2M & 64.2\0 & 82.2\0 & 37.7\0 & 54.8\0 \\
MCB+Att+GloVe~\citep{Fukui2016} & 70.5M & 64.7\0 & 82.5\0 & 37.6\0 & 55.6\0 \\
MCB+Att+Glove+VG~\citep{Fukui2016} & 70.5M & 65.4\0 & 82.3\0 & 37.2\0 & 57.4\0 \\
\end{tabular}
\end{center}
\end{table}

In this section, we conduct six experiments to select the proposed model, Multimodal Low-rank Bilinear Attention Networks (MLB). Each experiment controls other factors except one factor to assess the effect on accuracies. Based on MRN~\citep{Kim2016b}, we start our assessments with an initial option of $G=1$ and shortcut connections of MRN, called as Multimodal Attention Residual Networks (MARN). Notice that we use one embeddings for each visual feature for better performance, based on our preliminary experiment (not shown). We attribute this choice to the attention mechanism for visual features, which provides more capacity to learn visual features. We use the same hyper-parameters of MRN~\citep{Kim2016b}, without any explicit mention of this.

The VQA dataset~\citep{Antol2015} is used as a primary dataset, and, for data augmentation, question-answering annotations of Visual Genome~\citep{krishnavisualgenome} are used. Validation is performed on the VQA \textit{test-dev} split, and model comparison is based on the results of the VQA \textit{test-standard} split. For the comprehensive reviews of VQA tasks, please refer to \citet{Wu2016b} and \citet{Kafle2016}. The details about preprocessing, question and vision embedding, and hyperparameters used in our experiments are described in Appendix~\ref{sec:experiment_details}. The source code for the experiments is available in Github repository\footnote{\url{https://github.com/jnhwkim/MulLowBiVQA}}.

\paragraph{Number of Learning Blocks}

\citet{Kim2016b} argue that three-block layered MRN shows the best performance among one to four-block layered models, taking advantage of residual learning. However, we speculate that an introduction of attention mechanism makes deep networks hard to optimize. Therefore, we explore the number of learning blocks of MARN, which have an attention mechanism using low-rank bilinear pooling.

\paragraph{Number of Glimpses}

\citet{Fukui2016} show that the attention mechanism of two glimpses was an optimal choice. In a similar way, we assess one, two, and four-glimpse models.

\paragraph{Non-Linearity}

We assess three options applying non-linearity on low-rank bilinear pooling, vanilla, before Hadamard product as in Equation~\ref{eq:non-linear}, and after Hadamard product as in Equation~\ref{eq:afterhadamard}.

\paragraph{Answer Sampling}

VQA~\citep{Antol2015} dataset has ten answers from unique persons for each question, while Visual Genome~\citep{krishnavisualgenome} dataset has a single answer for each question. Since difficult or ambiguous questions may have divided answers, the probabilistic sampling from the distribution of answers can be utilized to optimize for the multiple answers. An instance~\footnote{\url{https://github.com/akirafukui/vqa-mcb/blob/5fea8/train/multi_att_2_glove/vqa_data_provider_layer.py\#L130}} can be found in \citet{Fukui2016}. We simplify the procedure as follows: \begin{align}
  p(a_1) &= \begin{cases}
    |a_1| / \Sigma_i |a_i|, & \text{if } |a_1| \geq 3\\ 
    0,  & \text{otherwise} 
  \end{cases}\\
  p(a_0) &= 1 - p(a_1)
\end{align}
where $|a_i|$ denotes the number of unique answer $a_i$ in a set of multiple answers, $a_0$ denotes a mode, which is the most frequent answer, and $a_1$ denotes the secondly most frequent answer. We define the divided answers as having at least three answers which are the secondly frequent one, for the evaluation metric of VQA~\citep{Antol2015}, \begin{align}
   \text{accuracy}(a_k) &= \min\left(|a_k|/3, 1 \right).
\end{align}
The rate of the divided answers is approximately $16.40\%$, and only $0.23\%$ of questions have more than two divided answers in VQA dataset. We assume that it eases the difficulty of convergence without severe degradation of performance.

\paragraph{Shortcut Connection}

The contribution of shortcut connections for residual learning is explored based on the observation of the competitive performance of single-block layered model. Since the usefulness of shortcut connections is linked to the network depth~\citep{He2015}.

\paragraph{Data Augmentation}

The data augmentation with Visual Genome~\citep{krishnavisualgenome} question answer annotations is explored. Visual Genome~\citep{krishnavisualgenome} originally provides 1.7 Million visual question answer annotations. After aligning to VQA, the valid number of question-answering pairs for training is 837,298, which is for distinct 99,280 images.

\section{Results}
\label{sec:results}

The six experiments are conducted sequentially. Each experiment determines experimental variables one by one. Refer to Table~\ref{tab:results}, which has six sectors divided by mid-rules.

\subsection{Six Experiment Results}

\paragraph{Number of Learning Blocks}

Though, MRN~\citep{Kim2016b} has the three-block layered architecture, MARN shows the best performance with two-block layered models (63.92\%). For the multiple glimpse models in the next experiment, we choose one-block layered model for its simplicity to extend, and competitive performance (63.79\%).

\begin{table}[t]
\caption{The VQA \textit{test-standard} results to compare with state-of-the-art. Notice that these results are trained by provided VQA train and validation splits, without any data augmentation.}
\label{tab:standard}
\begin{center}
\begin{tabular}{l c c c c c c c c}
& \multicolumn{4}{c}{\bf Open-Ended} & \multicolumn{4}{c}{\bf MC } \\
\cmidrule{2-5}
\cmidrule{6-9}
{\bf MODEL} & {\bf ALL} & {\bf Y/N} & {\bf NUM} & {\bf ETC} & {\bf ALL}
\\ \hline \\
iBOWIMG~\citep{Zhou2015}
         & 55.89 & 76.76 & 34.98 & 42.62
         & 61.97 \\
DPPnet~\citep{Noh2015}
         & 57.36 & 80.28 & 36.92 & 42.24
         & 62.69 \\
Deeper LSTM+Normalized CNN~\citep{Antol2015}
         & 58.16 & 80.56 & 36.53 & 43.73 
         & 63.09 \\
SMem~\citep{Xu2016}
         & 58.24 & 80.80 & 37.53 & 43.48 
         & - \\
Ask Your Neurons~\citep{Malinowski2016}
         & 58.43 & 78.24 & 36.27 & 46.32 
         & - \\
SAN~\citep{Yang2015}
         & 58.85 & 79.11 & 36.41 & 46.42
         & - \\
D-NMN~\citep{Andreas2016}
         & 59.44 & 80.98 & 37.48 & 45.81
         & - \\
ACK~\citep{Wu2016}
         & 59.44 & 81.07 & 37.12 & 45.83
         & - \\
FDA~\citep{Ilievski2016}
         & 59.54 & 81.34 & 35.67 & 46.10
         & 64.18 \\
HYBRID~\citep{Kafle2016a}
         & 60.06 & 80.34 & 37.82 & 47.56
         & - \\
DMN+~\citep{Xiong2016}
         & 60.36 & 80.43 & 36.82 & 48.33
         & - \\
MRN~\citep{Kim2016b}
         & 61.84 & 82.39 & \bf{38.23} & 49.41
         & 66.33 \\
HieCoAtt~\citep{Lu2016}
         & 62.06 & 79.95 & 38.22 & 51.95
         & 66.07 \\
RAU~\citep{Noh2016}
         & 63.2\0 & 81.7\0 & 38.2\0 & 52.8\0
         & 67.3\0 \\
\midrule[0.4pt]
MLB (ours)
         & \bf{65.07} & \bf{84.02} & 37.90 & \bf{54.77} 
         & \bf{68.89} \\
\end{tabular}
\end{center}
\end{table}

\paragraph{Number of Glimpses}

Compared with the results of \citet{Fukui2016}, four-glimpse MARN (64.61\%) is better than other comparative models. However, for a parsimonious choice, two-glimpse MARN (64.53\%) is chosen for later experiments. We speculate that multiple glimpses are one of key factors for the competitive performance of MCB~\citep{Fukui2016}, based on a large margin in accuracy, compared with one-glimpse MARN (63.79\%).

\paragraph{Non-Linearity}

The results confirm that activation functions are useful to improve performances. Surprisingly, there is no empirical difference between two options, before-Hadamard product and after-Hadamard product. This result may build a bridge to relate with studies on multiplicative integration with recurrent neural networks~\citep{Wu2016a}.

\paragraph{Answer Sampling}

Sampled answers (64.80\%) result better performance than mode answers (64.53\%). It confirms that the distribution of answers from annotators can be used to improve the performance. However, the number of multiple answers is usually limited due to the cost of data collection.

\paragraph{Shortcut Connection}

Though, MRN~\citep{Kim2016b} effectively uses shortcut connections to improve model performance, one-block layered MARN shows better performance without the shortcut connection. In other words, the residual learning is not used in our proposed model, MLB. It seems that there is a trade-off between introducing attention mechanism and residual learning. We leave a careful study on this trade-off for future work.

\paragraph{Data Augmentation}

Data augmentation using Visual Genome~\citep{krishnavisualgenome} question answer annotations significantly improves the performance by 0.76\% in accuracy for VQA \textit{test-dev} split. Especially, the accuracy of \textit{others} (ETC)-type answers is notably improved from the data augmentation.

\subsection{Comparison with State-of-the-Art}

The comparison with other single models on VQA \textit{test-standard} is shown in Table~\ref{tab:standard}. The overall accuracy of our model is approximately 1.9\% above the next best model~\citep{Noh2016} on the \textit{Open-Ended} task of VQA. The major improvements are from \textit{yes-or-no} (Y/N) and \textit{others} (ETC)-type answers. In Table~\ref{tab:ensemble}, we also report the accuracy of our ensemble model to compare with other ensemble models on VQA \textit{test-standard}, which won 1st to 5th places in VQA Challenge 2016\footnote{http://visualqa.org/challenge.html}. We beat the previous state-of-the-art with a margin of 0.42\%.

\renewcommand*{\thefootnote}{\fnsymbol{footnote}}
\begin{table}[ht!]
\caption{The VQA \textit{test-standard} results for ensemble models to compare with state-of-the-art. For unpublished entries, their team names are used instead of their model names. Some of their figures are updated after the challenge.}
\label{tab:ensemble}
\begin{center}
\begin{tabular}{l c c c c c c c c}
& \multicolumn{4}{c}{\bf Open-Ended} & \multicolumn{4}{c}{\bf MC } \\
\cmidrule{2-5}
\cmidrule{6-9}
{\bf MODEL} & {\bf ALL} & {\bf Y/N} & {\bf NUM} & {\bf ETC} & {\bf ALL}
\\ \hline \\
RAU~\citep{Noh2016}
         & 64.12 & 83.33 & 38.02 & 53.37
         & 67.34 \\
MRN~\citep{Kim2016b}
         & 63.18 & 83.16 & 39.14 & 51.33
         & 67.54 \\
DLAIT (not published)
         & 64.83 & 83.23 & \textbf{40.80} & 54.32
         & 68.30 \\
Naver Labs (not published)
         & 64.79 & 83.31 & 38.70 & 54.79
         & 69.26 \\
MCB~\citep{Fukui2016}
         & 66.47 & 83.24 & 39.47 & \textbf{58.00}
         & 70.10 \\
\midrule[0.4pt]
MLB (ours)
         & \textbf{66.89} & \textbf{84.61} & 39.07 & 57.79 
         & \textbf{70.29} \\
\midrule[0.4pt]
Human~\citep{Antol2015} 
         & 83.30 & 95.77 & 83.39 & 72.67 
         & 91.54 \\
\end{tabular}
\end{center}
\end{table}
\renewcommand*{\thefootnote}{\arabic{footnote}}

\section{Related Works}

MRN~\citep{Kim2016b} proposes multimodal residual learning with Hadamard product of low-rank bilinear pooling. However, their utilization of low-rank bilinear pooling is limited to joint residual mapping function for multimodal residual learning. Higher-order Boltzmann Machines~\citep{Memisevic2007,Memisevic2010} use Hadamard product to capture the interactions of input, output, and hidden representations for energy function. \citet{Wu2016a} propose the recurrent neural networks using Hadamard product to integrate multiplicative interactions among hidden representations in the model. For details of these related works, please refer to Appendix~\ref{sec:related_works}.

Yet, compact bilinear pooling or multimodal compact bilinear pooling~\citep{Gao2015a,Fukui2016} is worth to discuss and carefully compare with our method. 

\subsection{Compact Bilinear Pooling}

Compact bilinear pooling~\citep{Gao2015a} approximates full bilinear pooling using a sampling-based computation, Tensor Sketch Projection~\citep{charikar2002,pham2013}: \begin{align}
  \Psi(x \otimes y,h,s) &= \Psi(x,h,s)*\Psi(y,h,s) \\
                        &= \FFT^{-1}(\FFT(\Psi(x,h,s)\circ\FFT(\Psi(y,h,s))
\end{align}
where $\otimes$ denotes outer product, $*$ denotes convolution, $\Psi(v,h,s)_{i} := \sum_{j:h_j = i} s_j \cdot v_j$,
$\FFT$ denotes Fast Fourier Transform, $d$ denotes an output dimension, $x,y,h,s \in \R^n$, $x$ and $y$ are inputs, and $h$ and $s$ are random variables. $h_i$ is sampled from $\{1,...,d\}$, and $s_i$ is sampled from $\{-1,1\}$, then, both random variables are fixed for further usage. Even if the dimensions of $x$ and $y$ are different from each other, it can be used for multimodal learning~\citep{Fukui2016}.

Similarly to Equation~\ref{eq:bilin}, compact bilinear pooling can be described as follows: \begin{align}
  f_i &= \vx^T\mathcal{W}_i\vy
\end{align}
where $\mathcal{W}_{ijk} = s_{ijk} w_{ijk}$ if $s_{ijk}$ is sampled from $\{-1,1\}$, $w_{ijk}$ is sampled from $\{\mP_{i1}, \mP_{i2}, \dots, \mP_{id}\}$, and the compact bilinear pooling is followed by a fully connected layer $\mP \in \R^{|\Omega| \times d}$. Then, this method can be formulated as a hashing trick~\citep{weinberger2009,Chen2015a} to share randomly chosen bilinear weights using $d$ parameters for a output value, in a way that a single parameter is shared by $NM/d$ bilinear terms in expectation, with the variance of $NM(d-1)/d^2$ (See Appendix~\ref{sec:cbp}). 

In comparison with our method, their method approximates a three-dimensional weight tensor in bilinear pooling with a two-dimensional matrix $\mP$, which is larger than the concatenation of three two-dimensional matrices for low-rank bilinear pooling. The ratio of the number of parameters for a single output to the total number of parameters for $|\Omega|$ outputs is $d/d|\Omega| = 1/|\Omega|$~\citep{Fukui2016}, vs. $d(N+M+1) / d(N+M+|\Omega|) = (N+M+1)/(N+M+|\Omega|) \approx 2/3$ (ours), since our method uses a three-way factorization. Hence, more parameters are allocated to each bilinear approximation than compact bilinear pooling does, effectively managing overall parameters guided by back-propagation algorithm.

MCB~\citep{Fukui2016}, which uses compact bilinear pooling for multimodal tasks, needs to set the dimension of output $d$ to 16K, to reduce the bias induced by the fixed random variables $h$ and $s$. As a result, the majority of model parameters (16K $\times$ 3K = 48M) are concentrated on the last fully connected layer, which makes a \textit{fan-out} structure. So, the total number of parameters of MCB is highly sensitive to the number of classes, which is approximately 69.2M for \textit{MCB+att}, and 70.5M for \textit{MCB+att+GloVe}. Yet, the total number of parameters of our proposed model (MLB) is 51.9M, which is more robust to the number of classes having $d$ = 1.2K, which has a similar role in model architecture.

\section{Conclusions}

We suggest a low-rank bilinear pooling method to replace compact bilinear pooling, which has a \textit{fan-out} structure, and needs complex computations. Low-rank bilinear pooling has a flexible structure using linear mapping and Hadamard product, and a better parsimonious property, compared with compact bilinear pooling. We achieve new state-of-the-art results on the VQA dataset using a similar architecture of \citet{Fukui2016}, replacing compact bilinear pooling with low-rank bilinear pooling. We believe our method could be applicable to other bilinear learning tasks.

\subsubsection*{Acknowledgments}
\small
The authors would like to thank Patrick Emaase for helpful comments and editing. Also, we are thankful to anonymous reviewers who provided comments to improve this paper. This work was supported by NAVER LABS Corp. \& NAVER Corp. and partly by the Korea government (IITP-R0126-16-1072-SW.StarLab, KEIT-10044009-HRI.MESSI, KEIT-10060086-RISF, ADD-UD130070ID-BMRR). The part of computing resources used in this study was generously shared by Standigm Inc.

\small
\newpage
\bibliography{kim2016schur}
\bibliographystyle{iclr2017_conference}

\setcounter{section}{0}
\renewcommand\theHsection{\Alph{section}}
\renewcommand\thesection{\Alph{section}}
\renewcommand\thesubsection{\thesection.\arabic{subsection}}

\textbf{\Large{Appendix}}

\section{Experiment Details}
\label{sec:experiment_details}
\subsection{Preprocessing}

We follow the preprocessing procedure of \citet{Kim2016b}. Here, we remark some details of it, and changes.

\subsubsection{Question Embedding}

The 90.45\% of questions for the 2K-most frequent answers are used. The vocabulary size of questions is 15,031. GRU~\citep{Cho2014a} is used for question embedding. Based on earlier studies~\citep{Noh2015,Kim2016b}, a word embedding matrix and a GRU are initialized with Skip-thought Vector pre-trained model~\citep{Kiros2015}. As a result, question vectors have 2,400 dimensions.

For efficient computation of variable-length questions, \citet{Kim2016a} is used for the GRU. Moreover, for regularization, Bayesian Dropout~\citep{Gal2015} which is implemented in \citet{Leonard2015a} is applied while training.

\subsection{Vision Embedding}

ResNet-152 networks~\citep{He2015} are used for feature extraction. The dimensionality of an input image is $3 \times 448 \times 448$. The outputs of the last convolution layer is used, which have $2,048 \times 14 \times 14$ dimensions. 

\subsection{Hyperparameters}

The hyperparameters used in MLB of Table~\ref{tab:standard} are described in Table~\ref{tab:hyperparams}. The batch size is 100, and the number of iterations is fixed to 250K. For data augmented models, a simplified early stopping is used, starting from 250K to 350K-iteration for every 25K iterations (250K, 275K, 300K, 325K, and 350K; at most five points) to avoid exhaustive submissions to VQA \textit{test-dev} evaluation server. RMSProp~\citep{Tieleman2012} is used for optimization.

Though, the size of joint embedding size $d$ is borrowed from \citet{Kim2016b}, a grid search on $d$ confirms this choice in our model as shown in Table~\ref{tab:joint_embeeding_size}.

\begin{table}[h!]
\caption{Hyperparameters used in MLB (single model in Table~\ref{tab:standard}).}
\label{tab:hyperparams}
\begin{center}
\begin{tabular}{c r l}
{\bf SYMBOL} & {\bf VALUE} & {\bf DESCRIPTION}
\\ \hline \\
$S$ & 14 & attention lattice size \\
$N$ & 2,400 & question embedding size \\
$M$ & 2,048 & channel size of extracted visual features \\
$d$ & 1,200  & joint embedding size \\
$G$ & 2 & number of glimpses \\
$|\Omega|$ & 2,000 & number of candidate answers \\
$\eta$ & 3e-4 & learning rate \\
$\lambda$ & 0.99997592083 & learning rate decay factor at every iteration \\
$p$ & 0.5 & dropout rate \\
$\theta$ & $\pm$10 & gradient clipping threshold 
\end{tabular}
\end{center}

\end{table}\begin{table}[t]
\caption{The effect of joint embedding size $d$.}
\label{tab:joint_embeeding_size}
\begin{center}
\begin{tabular}{r c c c c c c c}
& & \multicolumn{4}{c}{\bf Open-Ended} \\
\cmidrule{3-6}
{\bf $d$} & {\bf SIZE} & {\bf ALL} & {\bf Y/N} & {\bf NUM} & {\bf ETC}
\\ \hline \\
800  & 45.0M & 64.89 & 84.08 & 38.15 & 54.55 \\
1000 & 48.4M & 65.06 & \bf{84.18} & 38.01 & 54.85 \\
1200 & 51.9M & \bf{65.08} & 84.14 & \bf{38.21} & \bf{54.87} \\
1400 & 55.4M & 64.94 & 84.13 & 38.00 & 54.64 \\
1600 & 58.8M & 65.02 & 84.15 & 37.79 & 54.85 \\
\end{tabular}
\end{center}
\end{table}

\subsection{Model Schema}

\begin{figure}[ht!]
\begin{center}
\includegraphics[width=.35\linewidth]{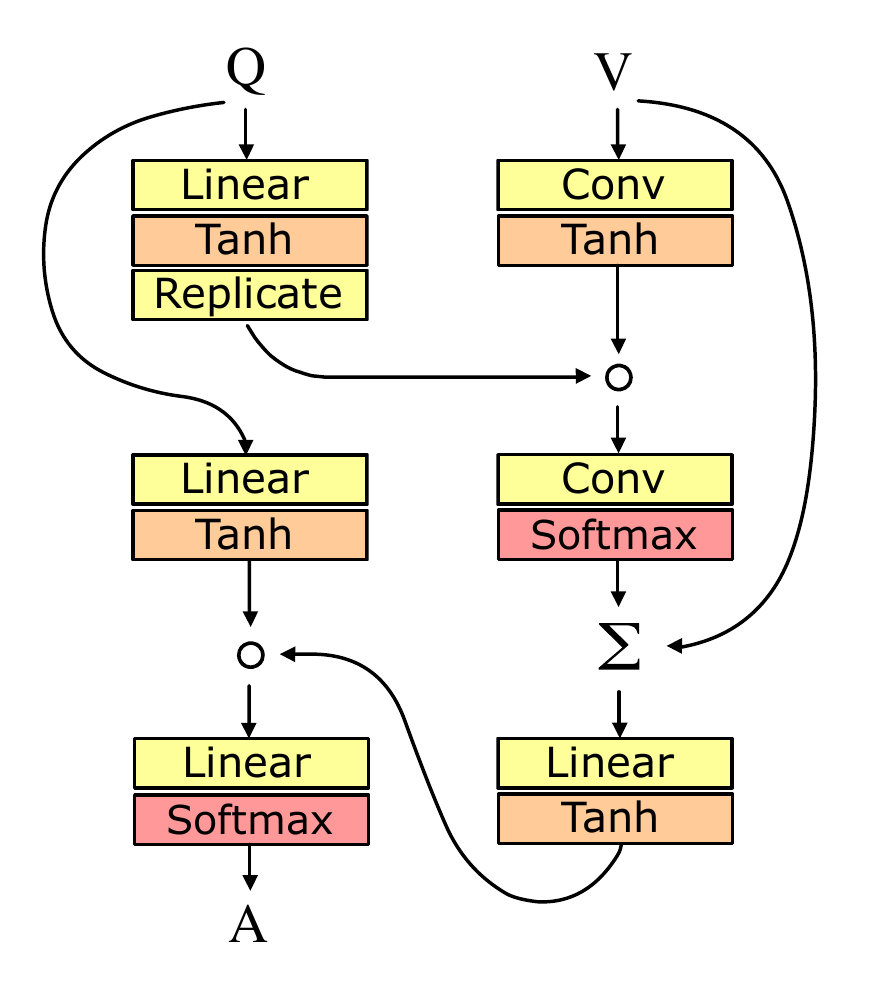}
\end{center}
\caption{A schematic diagram of MLB. \textit{Replicate} module copies an question embedding vector to match with $S^2$ visual feature vectors. \textit{Conv} modules indicate $1 \times 1$ convolution to transform a given channel space, which is computationally equivalent to linear projection for channels.}
\label{fig:schema}
\end{figure}

Figure~\ref{fig:schema} shows a schematic diagram of MLB, where $\circ$ denotes Hadamard product, and $\Sigma$ denotes a linear combination of visual feature vectors using coefficients, which is the output of softmax function. If $G > 1$, the softmax function is applied to each row vectors of an output matrix (Equation~\ref{eq:alpha}), and we concatenate the resulting vectors of the $G$ linear combinations (Equation~\ref{eq:concat}).

\subsection{Ensemble of Seven Models}

The \textit{test-dev} results for individual models consisting of our ensemble model is presented in Table~\ref{tab:ensemble7}. 

\begin{table}[t!]
\caption{The individual models used in our ensemble model in Table~\ref{tab:ensemble}.}
\label{tab:ensemble7}
\begin{center}
\begin{tabular}{l c c c c c c c}
& & \multicolumn{4}{c}{\bf Open-Ended} \\
\cmidrule{3-6}
{\bf MODEL} & {\bf GLIMPSE} & {\bf ALL} & {\bf Y/N} & {\bf NUM} & {\bf ETC}
\\ \hline \\
MLB    & 2 & 64.89 & 84.13 & 37.85 & 54.57 \\
MLB    & 2 & 65.08 & \bf{84.14} & 38.21 & 54.87 \\
MLB    & 4 & 65.01 & 84.09 & 37.66 & 54.88 \\
MLB-VG & 2 & 65.76 & 83.64 & 37.57 & 56.86 \\
MLB-VG & 2 & 65.84 & 83.87 & 37.87 & 56.76 \\
MLB-VG & 3 & 66.05 & 83.88 & 38.13 & 57.13 \\
MLB-VG & 4 & \bf{66.09} & 83.59 & \bf{38.32} & \bf{57.42} \\
\midrule[0.4pt]
Ensemble & - & 66.77 & 84.54 & 39.21 & 57.81 \\
\end{tabular}
\end{center}
\end{table}

\vspace*{2cm}
\vspace*{4\baselineskip}

\section{Understanding of Multimodal Compact Bilinear Pooling}
\label{sec:cbp}

In this section, the algorithm of multimodal compact bilinear pooling (MCB)~\citep{Gao2015a,Fukui2016} is described as a kind of hashing tick~\citep{Chen2015a}.

$\vx \in \R^{n_x}$ and $\vy \in \R^{n_y}$ are the given inputs, $\Phi(\vx,\vy) \in \R^d$ is the output. Random variables $\vh_x \in \N^{n_x}$ and $\vh_y \in \N^{n_y}$ are uniformly sampled from $\{1,\dots,d\}$, and $\vs_x \in \Z^{n_x}$ and $\vs_y \in \Z^{n_y}$ are uniformly sampled from $\{-1,1\}$. Then, Count Sketch projection function $\Psi$~\citep{charikar2002} projects $\vx$ and $\vy$ to intermediate representations $\Psi(\vx,\vh_x,\vs_x) \in \R^d$ and $\Psi(\vy,\vh_y,\vs_y) \in \R^d$, which is defined as: \begin{align}
   \Psi(\vv,\vh,\vs)_{i} := \sum_{j:h_j = i} s_j \cdot v_j
\end{align}
Notice that both $\vh$ and $\vs$ remain as constants after initialization~\citep{Fukui2016}.

The probability of $h_{xj}=i$ and $h_{yj}=i$ for the given $j$ is $1/d^2$. Hence, the expected number of bilinear terms in $\Psi(\vx,\vh_x,\vs_x)_i \Psi(\vy,\vh_y,\vs_y)_i$ is $(n_x n_y)/d^2$. Since, the output $\Phi(\vx,\vy)$ is a result of circular convolution of $\Psi(\vx,\vh_x,\vs_x)$ and $\Psi(\vy,\vh_y,\vs_y)$, the expected number of bilinear terms in $\Phi(\vx,\vy)_i$ is $(n_x n_y)/d$. Likewise, the probability of that a bilinear term is allocated in $\Phi(\vx, \vy)_i$ is $1/d$. The probability distribution of the number of bilinear terms in $\Phi(\vx, \vy)_i$ follows a multinomial distribution, whose mean is $(n_x n_y)/d$ and variance is $(n_x n_y)(d-1)/d^2$.

Linear projection after the multimodal compact bilinear pooling provides weights on the bilinear terms, in a way that a shared weight is assigned to $\Phi(\vx, \vy)_i$, which has $(n_x n_y)/d$ bilinear terms in expectation, though each bilinear term can have a different sign induced by both $\vs_x$ and $\vs_y$. 

HashedNets~\citep{Chen2015a} propose a method to compress neural networks using a low-cost hashing function~\citep{weinberger2009}, which is the same function of $\Psi(\vv,\vh,\vs)$. They randomly group a portion of connections in neural networks to share a single weight. We speculate that multimodal compact bilinear pooling uses the hashing tick to reduce the number of full bilinear weights with the rate of $d/(n_x n_y)$. However, this approximation is limited to two-way interaction, compared with three-way factorization in our method.

\section{Replacement of Low-rank Bilinear Pooling}

For the explicit comparison with compact bilinear pooling, we explicitly substitute compact bilinear pooling for low-rank bilinear pooling to control everything else, which means that the rest of the model architecture is exactly the same. 

According to \citet{Fukui2016}, we use MCB followed by Signed Square Root, L2-Normalization, Dropout ($p$=0.1), and linear projection from 16,000-dimension to the target dimension. Also, Dropout ($p$=0.3) for a question embedding vector. Note that an overall architecture for multimodal learning of both is the same. Experimental details are referenced from the implementation~\footnote{\url{https://github.com/akirafukui/vqa-mcb}} of \citet{Fukui2016}. 

For test-dev split, our version of MCB gets 61.48\% for overall accuracy (yes/no: 82.48\%, number: 37.06\%, and other: 49.07\%) vs. 65.08\% (ours, MLB in Table~\ref{tab:results}). Additionally, if the nonlinearity in getting attention distributions is increased as the original MCB does using ReLU, we get 62.11\% for overall accuracy (yes/no: 82.55\%, number: 37.18\%, and other: 50.30\%), which is still the below of our performance~\footnote{Our version of MCB definition can be found in \url{https://github.com/jnhwkim/MulLowBiVQA/blob/master/netdef/MCB.lua}}. 

We do not see it as a decisive evidence of the better performance of MLB, but as a reference (the comparison of test-dev results may be also unfair.), since an optimal architecture and hyperparameters may be required for each method.

\newpage
\section{Related Works}
\label{sec:related_works}

\subsection{Multimodal Residual Networks}
\label{subsec:mrn}


MRN~\citep{Kim2016b} is an implicit attentional model using multimodal residual learning with Hadamard product which does not have any explicit attention mechanism. \begin{align}
   \label{eq:general}
   \mathcal{F}^{(k)}(\vq,\vv) &= \sigma(\mW_{\vq}^{(k)} \vq) \circ \sigma(\mW_{2}^{(k)} \sigma(\mW_{1}^{(k)} \vv)) \\
   H_{L}(\vq,\vv) &= \mW_{\vq'} \vq + \sum_{l=1}^{L} \mW_{\mathcal{F}^{(l)}} \mathcal{F}^{(l)}(H_{l-1},\vv)
\end{align}
where $\mW_*$ are parameter matrices, $L$ is the number of learning blocks, $H_{0} = \vq$, $\mW_{\vq'} = \Pi_{l=1}^{L}\mW_{\vq'}^{(l)}$, and $\mW_{\mathcal{F}^{(l)}} = \Pi_{m=l+1}^{L}\mW_{\vq'}^{(m)}$. Notice that these equations can be generalized by Equation~\ref{eq:sc}.

However, an explicit attention mechanism allows the use of lower-level visual features than fully-connected layers, and, more importantly, spatially selective learning. Recent state-of-the-art methods use a variant of an explicit attention mechanism in their models~\citep{Lu2016,Noh2016,Fukui2016}. Note that shortcut connections of MRN are not used in the proposed Multimodal Low-rank Bilinear (MLB) model. Since, it does not have any performance gain due to not stacking multiple layers in MLB. We leave the study of residual learning for MLB for future work, which may leverage the excellency of bilinear models as suggested in \citet{Wu2016b}.

\subsection{Higher-Order Boltzmann Machines}

A similar model can be found in a study of Higher-Order Boltzmann Machines~\citep{Memisevic2007,Memisevic2010}. They suggest a factoring method for the three-way energy function to capture correlations among input, output, and hidden representations. \begin{align}
   -E(\vy,\vh;\vx) &= \sum_f\big(\sum_i x_iw_{if}^x\big)\big(\sum_j y_jw_{jf}^y\big)\big(\sum_k h_k w_{kf}^h\big)+\sum_k w_k^h h_k + \sum_j w_j^y y_j \nonumber \\
             &= \big(\vx^T\mW^x \circ \vy^T\mW^y \circ \vh^T\mW^h\big)\mathds{1} + \vh^T\vw^h + \vy^T \vw^y
\end{align}
Setting aside of bias terms, the $I \times J \times K$ parameter tensor of unfactored Higher-Order Boltzmann Machines is replaced with three matrices, $\mW^x \in \R^{I \times F}$, $\mW^y \in \R^{J \times F}$, and $\mW^h \in \R^{K \times F}$.

\subsection{Multiplicative Integration with Recurrent Neural Networks}

Most of recurrent neural networks, including vanilla RNNs, Long Short Term Memory networks~\citep{Hochreiter1997} and Gated Recurrent Units~\citep{Cho2014a}, share a common expression as follows: \begin{align}
  \phi(\mW\vx + \mU\vh + \vb)
\end{align}
where $\phi$ is a non-linear function, $\mW \in \R^{d \times n}$, $\vx \in \R^n$, $\mU \in \R^{d \times m}$, $\vh \in \R^m$, and $\vb \in \R^d$ is a bias vector. Note that, usually, $\vx$ is an input state vector and $\vh$ is an hidden state vector in recurrent neural networks.

\citet{Wu2016a} propose a new design to replace the additive expression with a multiplicative expression using Hadamard product as \begin{align}
  \phi(\mW\vx \circ \mU\vh + \vb).
\end{align}

Moreover, a general formulation of this multiplicative integration can be described as \begin{align}
  \phi(\pmb{\alpha} \circ \mW\vx \circ \mU\vh + \mW\vx \circ \pmb{\beta}_1 + \mU\vh \circ \pmb{\beta}_2 + \vb)
\end{align}
which is reminiscent of \textit{full model} in Section~\ref{sec:full}.

\end{document}